\documentclass{article} 
\usepackage{iclr2016_conference,times}
\usepackage{hyperref}
\usepackage{url}
\usepackage{graphicx}
\usepackage{subcaption}
\usepackage{import}
\usepackage{algorithm}
\usepackage{algorithmic}
\usepackage{amsfonts}
\usepackage{mathtools}
\usepackage{array}
\usepackage{longtable}

\graphicspath{{fig/}}

\title{Continuous control with deep reinforcement learning}

\author{
\hspace{-1mm}Timothy P.~Lillicrap\thanks{These authors contributed equally.},
Jonathan J.~Hunt\footnotemark[1],
Alexander Pritzel,
Nicolas Heess,\\
\textbf{Tom Erez,
Yuval Tassa,
David Silver \&
Daan Wierstra}
\\
Google Deepmind \\
London, UK\\
\texttt{\{countzero, jjhunt, apritzel, heess,
  } \\
  \texttt{\phantom{\{}etom, tassa, davidsilver,
  wierstra\} @ google.com}
}

\iclrfinalcopy 

\newcommand{\RR}{I\!\!R} 
\DeclareMathOperator{\E}{\mathbb{E}} 

\DeclareMathOperator*{\argmax}{arg\,max}

\begin{document}

\maketitle

\begin{abstract}
  We adapt the ideas underlying the success of Deep Q-Learning
  to the
  continuous action domain. We present an actor-critic, model-free
  algorithm based on the deterministic policy gradient that can operate
  over continuous action spaces.  Using the same learning algorithm,
  network architecture and hyper-parameters, our algorithm
  robustly solves more than 20 simulated physics tasks, including classic
  problems such as cartpole swing-up, dexterous manipulation,
  legged locomotion and car driving. Our algorithm is
  able to find policies whose performance is competitive with those
  found by a planning algorithm with full access to the dynamics
  of the domain and its derivatives. We further
  demonstrate that for many of the tasks the algorithm can learn
  policies ``end-to-end'': directly from raw pixel inputs.
\end{abstract}

\section{Introduction}

One of the primary goals of the field of artificial intelligence is to
solve complex tasks from
unprocessed, high-dimensional, sensory input.
Recently, significant progress has been made by combining advances in
deep learning for sensory processing \citep{krizhevsky2012imagenet}
with reinforcement learning,
resulting in the ``Deep Q Network'' (DQN) algorithm \citep{mnih2015human}
that is capable of human level performance on many Atari video games
using unprocessed pixels for input. To do so, deep neural
network function approximators were used to estimate the action-value
function.

However, while DQN solves problems with high-dimensional observation spaces,
it can only handle discrete and low-dimensional action spaces.
Many tasks of interest, most notably physical control tasks, have
continuous (real valued) and high dimensional action spaces.  DQN
cannot be straightforwardly applied
to continuous domains since it relies on a finding the
action that maximizes the action-value function, which in the
continuous valued case requires an iterative optimization process at
every step.


An obvious approach to adapting deep reinforcement learning
methods such as DQN to continuous
domains is to to simply discretize the action space. However, this
has many limitations, most notably the curse of dimensionality:
the number of actions increases exponentially
with the number of degrees of freedom. For example, a 7 degree of
freedom system (as in the human arm) with the coarsest discretization
$a_{i} \in \left\{ -k, 0, k \right\}$ for each joint leads to an
action space with dimensionality: $3^7 = 2187$.  The situation is even
worse for tasks that require fine control of actions as they require
a correspondingly finer grained discretization, leading to an
explosion of the number of discrete actions. Such large action spaces
are difficult to explore efficiently, and thus successfully training
DQN-like networks in this context is likely intractable.
Additionally, naive discretization of action spaces needlessly throws
away information about the structure of the action domain, which may be
essential for solving many problems.

In this work we present a model-free, off-policy actor-critic
algorithm using deep function approximators that can learn policies
in high-dimensional, continuous action spaces.  Our work is
based on the
deterministic policy gradient (DPG)
algorithm \citep{silver2014deterministic} (itself similar to NFQCA
\citep{hafner2011reinforcement}, and similar ideas can be found in
\citep{prokhorov1997adaptive}). However, as we show below,
a naive application of this actor-critic method with
neural function approximators is unstable for challenging problems.

Here we combine the actor-critic approach with insights from the
recent success of Deep Q Network (DQN) \citep{mnih2013playing,
  mnih2015human}.  Prior to DQN, it was generally believed that
learning value functions using large, non-linear function
approximators was difficult and unstable.  DQN is able to learn value
functions using such function approximators in a stable and robust way
due to two innovations: 1.\ the network is trained off-policy with
samples from a replay buffer to minimize correlations between samples;
2.\ the network is trained with a ‘target’ Q network to give
consistent targets during temporal difference backups.  In this work
we make use of the same ideas, along with batch normalization
\citep{ioffe2015batch}, a recent advance in deep learning.


In order to evaluate our method we constructed a variety of challenging physical control problems
that involve complex multi-joint movements, unstable and rich contact
dynamics, and gait behavior.  Among these are classic
problems such as the cartpole swing-up problem, as well as many new
domains.
A long-standing
challenge of robotic control is to learn an action policy directly from raw
sensory input such as video. Accordingly, we place a fixed viewpoint
camera in the simulator and attempted all tasks
using both low-dimensional observations (e.g.\ joint angles)
and directly from pixels.

Our model-free approach which we call
Deep DPG (DDPG) can learn competitive policies for
all of our tasks
using low-dimensional observations
(e.g.\ cartesian coordinates or
joint angles) using the same hyper-parameters and
network structure.
In many cases, we are also able to learn good policies directly
from pixels, again keeping hyperparameters and
network structure constant \footnote{You can view a movie of some
of the learned policies at \url{https://goo.gl/J4PIAz}}.

A key feature of the approach is its simplicity: it requires
only a straightforward actor-critic architecture and learning
algorithm with very few ``moving parts'', making it easy to implement and
scale to more difficult problems and larger networks.  For the
physical control problems we compare our results to a baseline
computed by a planner \citep{tassa2012synthesis} that has full access
to the underlying simulated dynamics and its derivatives (see
supplementary information).  Interestingly, DDPG can sometimes find
policies that exceed the performance of the planner, in some cases
even when learning from pixels (the planner always plans over the
underlying low-dimensional state space).


\section{Background}

We consider a standard reinforcement learning setup consisting of an
agent interacting with an environment $E$ in discrete timesteps. At
each timestep $t$ the agent receives an observation $x_t$,
takes an action $a_t $ and receives a scalar reward $r_t$.
In all the environments considered here the actions are real-valued
$a_t \in \RR^N $.
In general, the environment may be
partially observed so that the entire history of the observation, action
pairs $ s_t = (x_1, a_1, ... , a_{t-1}, x_t) $
may be required to describe the state. Here, we
assumed the environment is fully-observed so
 $s_t = x_t$.

An agent's behavior is defined by a policy, $\pi$, which maps states
to a probability distribution over the actions $ \pi \colon
\mathcal{S} \to \mathcal{P(A)} $. The environment, $E$, may also be
stochastic.  We model it as a Markov decision process with a state
space $\mathcal{S}$, action space $\mathcal{A} = \RR^N$, an initial
state distribution $p(s_1)$, transition dynamics $p(s_{t+1} | s_t,
a_t)$, and reward function $r(s_t, a_t)$.

The return from a state is defined as the sum of discounted future reward
$R_t = \sum_{i = t}^{T} \gamma^{(i - t)} r(s_i, a_i) $
with a discounting factor
$\gamma \in [0, 1] $. Note that the return depends on the actions
chosen, and therefore on the policy $ \pi $, and may be stochastic.
The goal in reinforcement learning is
to learn a policy which maximizes the expected return from the start
distribution $J = \E_{r_i, s_i \sim E, a_i \sim \pi} \left[ R_1 \right ]$. We
denote the discounted state
visitation distribution for a policy $\pi$ as $\rho^\pi$.

The action-value function is used in many reinforcement learning algorithms. It
describes the expected return after taking an action $a_t$ in state $s_t$ and
thereafter following policy $\pi$:
\begin{equation}
  Q^\pi(s_t, a_t) = \E_{r_{i \ge t}, s_{i > t} \sim E, a_{i > t} \sim \pi} \left[ R_t | s_t, a_t \right ]
\end{equation}

Many approaches in reinforcement learning make use of the recursive
relationship known as the Bellman equation:
\begin{equation} \label{eq:bellndet}
  Q^{\pi}(s_t, a_t) = \E_{r_t, s_{t+1} \sim E} \left[ r(s_t, a_t) +
    \gamma \E_{a_{t + 1} \sim \pi} \left[
      Q^{\pi}(s_{t+1}, a_{t + 1}) \right]  \right]
\end{equation}

If the target policy is deterministic we can describe it as a function
$\mu: \mathcal{S} \leftarrow \mathcal{A}$ and avoid the inner expectation:
\begin{equation} \label{eq:bell}
  Q^{\mu}(s_t, a_t) = \E_{r_t, s_{t+1} \sim E} \left [ r(s_t, a_t) +
    \gamma
      Q^{\mu}(s_{t+1}, \mu(s_{t + 1})) \right]
\end{equation}

The expectation depends only on the environment.
This means that it is possible to learn $Q^{\mu}$ off-policy, using
transitions which are generated from a different stochastic
behavior policy $\beta$.

Q-learning \citep{watkins1992q}, a commonly used off-policy algorithm,
uses the greedy policy $\mu(s) = \argmax_a Q(s, a)$. We consider
function approximators parameterized by $\theta^Q$, which we optimize
by minimizing the loss:
\begin{equation} \label{eq:dqnloss}
  L(\theta^{Q}) = \E_{s_t \sim \rho^\beta, a_t \sim \beta, r_t \sim E}
  \left[\left(Q(s_t, a_t | \theta^{Q}) - y_t\right)^2 \right]
\end{equation}
where
\begin{equation} \label{eq:dqn}
  y_t =  r(s_t, a_t) +
      \gamma Q(s_{t+1}, \mu(s_{t+1}) | \theta^Q).
\end{equation}
While $y_t$ is also dependent on $\theta^Q$, this is typically ignored.

The use of large, non-linear function approximators for learning value
or action-value functions has often been avoided in the past since
theoretical performance guarantees are impossible, and practically
learning tends to be unstable.  Recently,
\citep{mnih2013playing,mnih2015human} adapted the Q-learning algorithm
in order to make effective use of large neural networks as function
approximators.  Their algorithm was able to learn to play Atari games
from pixels.  In order to scale Q-learning they introduced two major
changes: the use of a \textit{replay buffer}, and a separate
\textit{target network} for calculating $y_t$. We employ these in the
context of DDPG and explain their implementation in the next section.

\section{Algorithm}

It is not possible to straightforwardly apply Q-learning to continuous
action spaces, because in continuous spaces finding the greedy policy requires an
optimization of $a_t$ at every
timestep; this optimization is too slow to be practical with large, unconstrained
function approximators and nontrivial action spaces.
Instead, here we used an actor-critic approach based
on the DPG algorithm \citep{silver2014deterministic}.

The DPG algorithm
maintains a parameterized actor function $\mu(s | \theta^\mu)$
which specifies the current policy by deterministically mapping states to a
specific action. The critic $Q(s, a)$ is learned using the
Bellman equation as in Q-learning. The actor is updated by following the
applying the chain rule to the expected return from the start distribution
$J$ with respect to the actor parameters:
\begin{equation}
  \begin{split}
  \nabla_{\theta^\mu} J &\approx
  \E_{s_t \sim \rho^\beta}\left[\nabla_{\theta^{\mu}}
    Q(s, a | \theta^Q)|_{s = s_t, a = \mu(s_t | \theta^{\mu})}
                   \right] \\
                 & =
    \E_{s_t \sim \rho^\beta}\left[\nabla_{a} Q(s, a | \theta^{Q})|_{s = s_t, a = \mu(s_t)}
    \nabla_{\theta_\mu} \mu(s | \theta^{\mu})|_{s = s_t} \right]
  \end{split}
\end{equation}
\citet{silver2014deterministic} proved that this is
the \emph{policy gradient}, the gradient of the policy's
performance \footnote{In practice, as in commonly done in policy gradient implementations,
we ignored the discount in the state-visitation distribution $\rho^\beta$.}.

As with Q learning, introducing non-linear function approximators
means that convergence is no longer guaranteed. However, such
approximators appear essential in order to learn and generalize on
large state spaces. NFQCA \citep{hafner2011reinforcement}, which uses
the same update rules as DPG but with neural network function
approximators, uses batch learning for stability, which is intractable
for large networks.
A minibatch version of NFQCA which does not reset
the policy at each update, as would be required to scale to large networks,
is equivalent to the original DPG, which we compare to here.
Our contribution here is to provide
modifications to DPG, inspired by the success of DQN, which allow it
to use neural network function approximators to learn in
large state and action spaces online. We refer to our algorithm as
Deep DPG (DDPG, Algorithm \ref{algo:ddpg}).

One challenge when using neural networks for reinforcement learning is
that most optimization algorithms assume that the samples are independently
and identically distributed. Obviously, when the samples are generated
from exploring sequentially in an environment this assumption no longer holds.
Additionally, to make efficient use of hardware optimizations, it is essential
to learn in minibatches, rather than online.

As in DQN, we used a replay buffer to address
these issues. The replay buffer is a finite sized cache $\mathcal{R}$.
Transitions were sampled
from the environment according to the exploration policy and the tuple
$(s_t, a_t, r_t, s_{t + 1})$ was stored in the replay buffer.
When the replay buffer was full the oldest samples were discarded.
At each timestep the actor and critic are updated by sampling a minibatch
uniformly from the buffer. Because DDPG is an off-policy algorithm, the
replay buffer can be large, allowing the algorithm to benefit from
learning across a set of uncorrelated transitions.

Directly implementing Q learning (equation \ref{eq:dqnloss}) with neural
networks proved to be unstable in many environments. Since the network
$Q(s, a | \theta^Q)$ being updated is also used in calculating the target value
(equation \ref{eq:dqn}),
the Q update is prone to divergence.
Our solution is similar to the
target network used in \citep{mnih2013playing} but modified for actor-critic
and using ``soft'' target updates, rather than directly copying the
weights. We
create a copy of the actor and critic networks, $Q'(s, a | \theta^{Q'})$
and $\mu'(s | \theta^{\mu'})$ respectively, that are used for calculating
the target values.
The weights of these target networks
are then updated by having them slowly track the learned networks:
$\theta' \leftarrow \tau\theta + (1 - \tau) \theta'$ with $\tau \ll 1$.
This means that the
target values are constrained to change slowly, greatly improving the
stability of learning. This simple
change moves the relatively unstable problem of
learning the action-value function closer to the case of
supervised learning,
a problem for which robust solutions exist.
We found that having both a target $\mu'$ and $Q'$ was required
to have stable targets $y_i$ in order to consistently
train the critic without divergence.
This may slow learning,
since the target network delays the propagation of value estimations. However,
in practice we found this was greatly outweighed by the stability of learning.

When learning from low dimensional feature vector observations, the different
components of the observation may have different physical units (for
example, positions versus velocities) and the ranges may vary
across environments. This can make it difficult for
the network to learn effectively and may make it difficult to find
hyper-parameters which generalise across environments with different
scales of state values.

One approach to
this problem is to manually scale the features so they are in similar
ranges across environments and units.
We address this issue by adapting a recent
technique from deep learning called \emph{batch normalization}
\citep{ioffe2015batch}. This technique normalizes each dimension
across the samples in a minibatch to have unit mean and variance.
In addition, it maintains a running average of the mean and variance to
use for normalization during testing
(in our case, during exploration or evaluation).
In deep networks, it is used to minimize covariance shift during training,
by ensuring that each layer receives whitened input.
In the low-dimensional case, we used batch
normalization on the state input and all layers of the  $\mu$
network and all layers of the $Q$ network prior to the action input
(details of the networks are given in the supplementary material).
With batch normalization, we were able
to learn effectively across many different tasks with differing types
of units, without needing to manually ensure the units were within a
set range.

A major challenge of learning in continuous action spaces is exploration.
An advantage of off-policies algorithms such
as DDPG is that we can treat the problem of exploration independently
from the learning algorithm. We constructed an exploration policy
$\mu'$ by adding noise
sampled from a noise process $\mathcal{N}$ to our actor policy
\begin{equation}
  \mu'(s_t) = \mu(s_t | \theta^\mu_t) + \mathcal{N}
\end{equation}
$\mathcal{N}$ can be chosen to suit the environment.  As
detailed in the supplementary materials we used an Ornstein–-Uhlenbeck
process \citep{uhlenbeck1930theory} to generate temporally correlated
exploration for exploration efficiency in physical control problems
with inertia (similar use of autocorrelated noise was introduced in
\citep{wawrzynski2015control}).

\begin{algorithm}[h]
  \caption{DDPG algorithm \label{algo:ddpg}}
  \label{dpgalgo}
  \begin{algorithmic}
    \STATE Randomly initialize critic network $Q(s, a | \theta^Q)$ and actor
    $\mu(s | \theta^{\mu})$ with weights $\theta^{Q}$ and $\theta^{\mu}$.
    \STATE Initialize target network $Q'$ and $\mu'$ with weights $\theta^{Q'}
    \leftarrow \theta^{Q}$, $\theta^{\mu'} \leftarrow \theta^{\mu}$
    \STATE Initialize replay buffer $R$
    \FOR{episode = 1, M}
      \STATE Initialize a random process $\mathcal{N}$ for action
      exploration
      \STATE Receive initial observation state $s_1$
      \FOR{t = 1, T}
        \STATE Select action $a_t = \mu(s_t | \theta^{\mu}) + \mathcal{N}_t$
        according to the current policy and exploration noise
        \STATE Execute action $a_t$ and observe
        reward $r_t$ and observe new state $s_{t+1}$
        \STATE Store transition $(s_t, a_t,
                r_t, s_{t+1})$ in $R$
        \STATE Sample a random minibatch of $N$ transitions
               $(s_i, a_i,
        r_i, s_{i + 1})$ from $R$
        \STATE Set $ y_i = r_i + \gamma Q'(s_{i + 1},
        \mu'(s_{i+1} | \theta^{\mu'}) | \theta^{Q'}) $
        \STATE Update critic by minimizing the loss:
               $L = \frac{1}{N} \sum_i (y_i -
               Q(s_i, a_i | \theta^Q))^2$
        \STATE Update the actor policy using the sampled policy gradient:
        \begin{equation*}
            \nabla_{\theta^{\mu}} J \approx
            \frac{1}{N} \sum_i
               \nabla_{a} Q(s, a | \theta^Q)|_{s = s_i, a = \mu(s_i)}
               \nabla_{\theta^\mu} \mu(s | \theta^\mu)|_{s_i}
         \end{equation*}
        \STATE Update the target networks:
          \begin{equation*}
            \theta^{Q'} \leftarrow \tau \theta^{Q} + (1 - \tau) \theta^{Q'}
          \end{equation*}
          \begin{equation*}
            \theta^{\mu'} \leftarrow \tau \theta^{\mu} +
                (1 - \tau) \theta^{\mu'}
          \end{equation*}
        \ENDFOR
    \ENDFOR
  \end{algorithmic}
\end{algorithm}

\section{Results}

\begin{figure}[h]
\begin{center}
  \includegraphics[width=\linewidth]{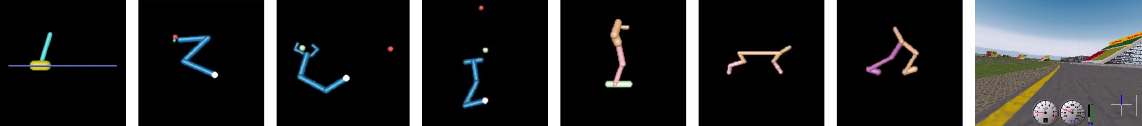}
\end{center}
\caption{ Example screenshots of a sample of environments we attempt
  to solve with DDPG. In order from the left: the cartpole swing-up
  task, a reaching task, a gasp and move task, a puck-hitting task, a
  monoped balancing task, two locomotion tasks and Torcs (driving
  simulator). We tackle all tasks using both low-dimensional feature
  vector and high-dimensional pixel inputs. Detailed descriptions of
  the environments are provided in the supplementary. Movies of some of
  the learned policies are available at \url{https://goo.gl/J4PIAz}.
  }
\label{fig:environments}
\end{figure}

We constructed simulated physical environments of varying levels of
difficulty to test our algorithm. This included classic reinforcement
learning environments such as cartpole, as well as difficult, high
dimensional tasks such as gripper, tasks involving contacts such as
puck striking (\emph{canada}) and locomotion tasks such as
\emph{cheetah} \citep{wawrzynski2009real}.  In all domains but
\emph{cheetah} the actions were torques applied to the actuated joints.
These environments were simulated using
MuJoCo \citep{todorov2012mujoco}.
Figure \ref{fig:environments} shows renderings of
some of the environments used in the task (the supplementary contains
details of the environments and you can view some of the learned policies at
\url{https://goo.gl/J4PIAz}).


In all tasks, we ran experiments using both a low-dimensional state
description (such as joint angles and positions) and high-dimensional
renderings of the environment. As in DQN \citep{mnih2013playing,
  mnih2015human}, in order to make the problems approximately fully
observable in the high dimensional environment we used action
repeats. For each timestep of the agent, we step the simulation 3
timesteps, repeating the agent's action and rendering each time.  Thus
the observation reported to the agent contains 9 feature maps (the RGB
of each of the 3 renderings) which allows the agent to infer
velocities using the differences between frames. The frames were
downsampled to 64x64 pixels and the 8-bit RGB values were converted to
floating point scaled to $[0, 1]$. See supplementary information for
details of our network structure and hyperparameters.

\begin{figure}[h]
\begin{center}
  \includegraphics[width=\linewidth]{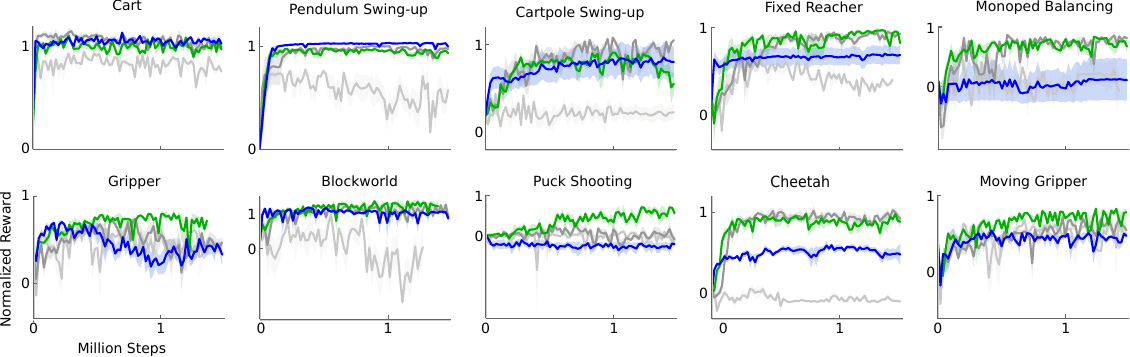}
\end{center}
\caption{ Performance curves for a selection of domains using variants of
  DPG: original DPG algorithm (minibatch NFQCA) with batch
  normalization (light grey), with target network (dark grey), with
  target networks and batch normalization (green), with target
  networks from pixel-only inputs (blue).  Target
  networks are crucial.  }
\label{fig:curves}
\end{figure}

We evaluated the policy periodically during training by testing it without
exploration noise. Figure
\ref{fig:curves} shows the performance curve for a selection of environments.
We also report results with components of our algorithm (i.e. the target
network or batch normalization) removed. In order to perform well across
all tasks, both of these additions are necessary. In particular, learning
without a target network, as in the original work with DPG,
is very poor in many environments.

Surprisingly, in some simpler tasks, learning policies from pixels is
just as fast as learning using the low-dimensional state
descriptor.  This may be due to the action repeats making the problem
simpler.  It may also be that the convolutional
layers provide an easily separable representation of state space,
which is straightforward for the higher layers to learn on quickly.

Table \ref{table:results} summarizes DDPG's performance across all of
the environments (results are averaged over 5
replicas).
We normalized the scores using two baselines.
The first baseline is the mean return from a naive policy
which samples actions from a uniform distribution over the valid
action space. The second baseline is iLQG
\citep{todorov2005generalized}, a planning based solver with full
access to the underlying physical model and its derivatives. We
normalize scores so that the naive policy has a mean score of $0$ and
iLQG has a mean score of $1$.
DDPG is
able to learn good policies on many of the tasks, and in many cases
some of the replicas learn policies which are superior to those found
by iLQG, even when learning directly from pixels.

\begin{figure}[h]
\begin{center}
  \includegraphics[width=0.7\linewidth]{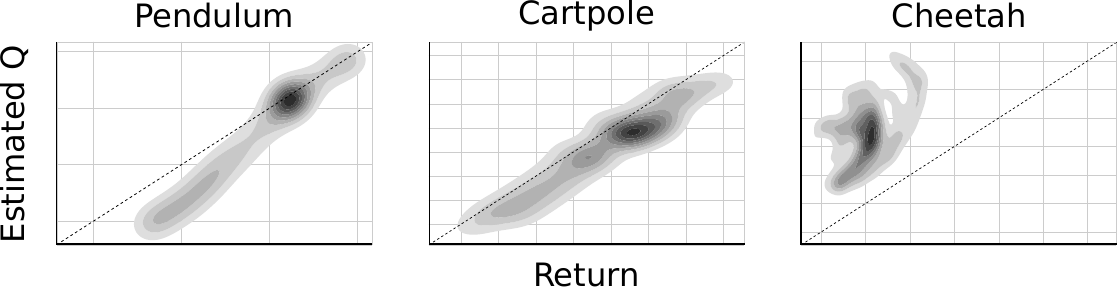}
\end{center}
\caption{ Density plot showing estimated Q values
  versus observed returns sampled from test episodes
  on 5 replicas.
  In simple domains such as
  pendulum and cartpole the Q values are
  quite accurate. In more complex tasks, the Q estimates are less accurate, but
  can still be used to learn competent policies. Dotted line
  indicates unity, units are arbitrary.
}
\label{fig:returns}
\end{figure}

It can be challenging to learn accurate value estimates. Q-learning,
for example, is prone to over-estimating values
\citep{hasselt2010double}.  We examined DDPG's estimates empirically by
comparing the values estimated by $Q$ after training with the true
returns seen on test episodes. Figure \ref{fig:returns} shows that in
simple tasks DDPG estimates returns accurately without systematic
biases. For harder tasks the Q estimates are worse, but DDPG is still
able learn good policies.



To demonstrate the generality of our approach
we also include Torcs, a racing game where the actions are acceleration,
braking
and steering. Torcs has previously been used as a testbed in other
policy learning approaches \citep{koutnik2014online}.
We used an identical network architecture and
learning algorithm hyper-parameters to the physics tasks but altered
the noise process for exploration because of the very different time
scales involved.  On both low-dimensional and from pixels, some
replicas were able to learn reasonable policies that are able to
complete a circuit around the track though other replicas failed to
learn a sensible policy.

\begin{table}[t]
  \caption{ Performance after training across all environments for at most 2.5
    million steps.  We report both the average and best observed
    (across 5 runs). All scores, except Torcs, are normalized so that
    a random agent receives 0 and a planning algorithm 1; for Torcs we
    present the raw reward score. We include results from the DDPG algorithn
    in the low-dimensional
    ($lowd$) version of the environment and high-dimensional ($pix$). For comparision
    we also include results from the original DPG algorithm with a replay
    buffer and batch normalization ($cntrl$).
  } \label{table:results}
\label{eval-table}
\begin{center}
  \begin{tabular}{ c | c | c | c | c | c | c }
    environment & $R_{av,lowd}$ & $R_{best,lowd}$ & $R_{av,pix}$
    & $R_{best,pix}$ & $R_{av,cntrl}$ & $R_{best,cntrl}$\\
    \hline
      blockworld1 & 1.156 & 1.511 & 0.466 & 1.299 &  -0.080 & 1.260 \\
      blockworld3da & 0.340 & 0.705 & 0.889 & 2.225 & -0.139 & 0.658 \\
      canada & 0.303 & 1.735 & 0.176 & 0.688 & 0.125 & 1.157 \\
      canada2d & 0.400 & 0.978 & -0.285 & 0.119 & -0.045 & 0.701 \\
      cart & 0.938 & 1.336 & 1.096 & 1.258 & 0.343 & 1.216 \\
      cartpole & 0.844 & 1.115 & 0.482 & 1.138 & 0.244 & 0.755 \\
      cartpoleBalance & 0.951 & 1.000 & 0.335 & 0.996 & -0.468 & 0.528 \\
      cartpoleParallelDouble & 0.549 & 0.900 & 0.188 & 0.323 & 0.197 & 0.572 \\
      cartpoleSerialDouble & 0.272 & 0.719 & 0.195 & 0.642 & 0.143 & 0.701 \\
      cartpoleSerialTriple & 0.736 & 0.946 & 0.412 & 0.427 & 0.583 & 0.942 \\
      cheetah & 0.903 & 1.206 & 0.457 & 0.792 & -0.008 & 0.425 \\
      fixedReacher & 0.849 & 1.021 & 0.693 & 0.981 & 0.259 & 0.927 \\
      fixedReacherDouble & 0.924 & 0.996 & 0.872 & 0.943 & 0.290 & 0.995 \\
      fixedReacherSingle & 0.954 & 1.000 & 0.827 & 0.995 & 0.620 & 0.999 \\
      gripper & 0.655 & 0.972 & 0.406 & 0.790 & 0.461 & 0.816 \\
      gripperRandom & 0.618 & 0.937 & 0.082 & 0.791 & 0.557 & 0.808 \\
      hardCheetah & 1.311 & 1.990 & 1.204 & 1.431 & -0.031 & 1.411 \\
      hopper & 0.676 & 0.936 & 0.112 & 0.924 & 0.078 & 0.917 \\
      hyq & 0.416 & 0.722 & 0.234 & 0.672 & 0.198 & 0.618 \\
      movingGripper & 0.474 & 0.936 & 0.480 & 0.644 & 0.416 & 0.805 \\
      pendulum & 0.946 & 1.021 & 0.663 & 1.055 & 0.099 & 0.951 \\
      reacher & 0.720 & 0.987 & 0.194 & 0.878 & 0.231 & 0.953 \\
      reacher3daFixedTarget & 0.585 & 0.943 & 0.453 & 0.922 & 0.204 & 0.631 \\
      reacher3daRandomTarget & 0.467 & 0.739 & 0.374 & 0.735 & -0.046 & 0.158 \\
      reacherSingle & 0.981 & 1.102 & 1.000 & 1.083 & 1.010 & 1.083 \\
      walker2d & 0.705 & 1.573 & 0.944 & 1.476 & 0.393 & 1.397 \\
      \hline
      torcs & -393.385 & 1840.036 & -401.911 & 1876.284 & -911.034 & 1961.600 \\
  \end{tabular}
\end{center}
\end{table}



\section{Related work}

The original DPG paper evaluated the algorithm with toy problems using
tile-coding and linear function approximators.  It demonstrated data
efficiency advantages for off-policy DPG over both on- and off-policy
stochastic actor critic.  It also solved one more challenging task in
which a multi-jointed octopus arm had to strike a target with any part
of the limb. However, that paper did not demonstrate scaling the
approach to large, high-dimensional observation spaces as we have
here.

It has often been assumed that standard policy search methods such as
those explored in the present work are simply too fragile to scale to
difficult problems \citep{levine2015end}.  Standard policy search is
thought to be difficult because it deals simultaneously with complex
environmental dynamics and a complex policy.
Indeed,
most past work with actor-critic and policy optimization
approaches have had difficulty scaling up to more
challenging problems \citep{deisenroth2013survey}.
Typically, this is due to instability in
learning wherein progress on a problem is either destroyed by subsequent
learning updates, or else learning is too slow to be practical.

Recent work with model-free policy search has
demonstrated that it may not be as fragile as previously supposed.
\citet{wawrzynski2009real, wawrzynski2013autonomous} has
trained stochastic policies in an actor-critic framework
with a replay buffer.
Concurrent with our work, \citet{balduzzi2015compatible} extended the DPG
algorithm with a ``deviator'' network which explicitly learns
$\partial Q/\partial a$. However, they only train on two low-dimensional domains.
\citet{heess2015learning} introduced SVG(0) which also uses
a Q-critic but learns a stochastic policy.
DPG can be considered the deterministic limit of SVG(0). The techniques we
described here for scaling DPG are also applicable to stochastic policies by
using the reparametrization trick \citep{heessRDPG2015,schulman2015gradient}.

Another approach, trust region policy
optimization (TRPO) \citep{schulman2015trust}, directly constructs
stochastic neural network policies without decomposing problems into
optimal control and supervised phases.  This method produces near
monotonic improvements in return by making carefully chosen updates to
the policy parameters, constraining updates to prevent the new policy
from diverging too far from the existing policy.  This approach does
not require learning an action-value function, and (perhaps as a
result) appears to be significantly less data efficient.

To combat the challenges of the actor-critic approach, recent work
with guided policy search (GPS) algorithms (e.g.,
\citep{levine2015end}) decomposes the problem into three phases that are
relatively easy to solve: first, it uses full-state observations to
create locally-linear approximations of the dynamics around one or
more nominal trajectories, and then uses optimal control to find the
locally-linear optimal policy along these trajectories; finally, it uses
supervised learning to train a complex, non-linear policy (e.g. a deep
neural network) to reproduce the state-to-action mapping of the optimized
trajectories.

This approach has several benefits, including data efficiency, and has been
applied successfully to a variety of real-world
robotic manipulation tasks using vision.
In these tasks GPS uses a similar convolutional policy network to ours with
2 notable exceptions: 1.\ it uses a spatial softmax to reduce the
dimensionality of visual features into a single $(x,y)$ coordinate for
each feature map, and
2.\ the policy also receives direct low-dimensional state information
about the configuration of the robot at the first fully connected
layer in the network.  Both likely increase the power and data
efficiency of the algorithm and could easily be exploited within the DDPG
framework.

PILCO \citep{deisenroth2011pilco} uses Gaussian processes to learn a
non-parametric, probabilistic model of the dynamics. Using this learned model,
PILCO calculates analytic policy gradients and achieves impressive
data efficiency in a number of control problems. However, due to the
high computational demand, PILCO is
``impractical for high-dimensional problems''
\citep{wahlstrom2015pixels}. It seems that deep function approximators
are the most promising approach for scaling reinforcement learning to
large, high-dimensional domains.

\citet{wahlstrom2015pixels}
used a deep dynamical model network
along with model predictive control to solve the pendulum
swing-up task from pixel input.  They trained a differentiable forward model
and encoded the goal state into the learned latent space.
They use model-predictive control over the learned model
to find a policy for reaching the target.
However, this approach is only applicable to domains with goal states that
can be demonstrated to the algorithm.

Recently, evolutionary approaches have been used to learn
competitive policies for Torcs from pixels
using compressed weight parametrizations \citep{koutnik2014evolving} or
unsupervised learning \citep{koutnik2014online}
to reduce the dimensionality
of the evolved weights. It is unclear how well these
approaches generalize to other problems.

\section{Conclusion}

The work combines insights from recent advances in deep
learning and reinforcement learning, resulting in an algorithm that
robustly solves challenging problems across a variety of domains with
continuous action spaces, even when using raw pixels for observations.
As with most reinforcement learning algorithms, the
use of non-linear function approximators nullifies any convergence
guarantees; however, our experimental results demonstrate that stable
learning without the need for any modifications between environments.

Interestingly, all of our experiments used substantially fewer steps
of experience than was used by DQN learning to find solutions in the
Atari domain.  Nearly all of the problems we looked at were solved
within 2.5 million steps of experience (and usually far fewer), a
factor of 20 fewer steps than DQN requires for good Atari solutions.
This suggests that, given more simulation time, DDPG may solve
even more difficult problems than those considered here.

A few limitations to our approach remain. Most notably, as with most
model-free reinforcement approaches, DDPG requires a large number of
training episodes to find solutions. However, we believe that a robust
model-free approach may be an important component of larger systems
which may attack these limitations \citep{glascher2010states}.

%


\bibliography{nndpg}
\bibliographystyle{iclr2016_conference}

\newpage

\begin{center}
{\LARGE\bf Supplementary Information: Continuous control with deep
  reinforcement learning}
\end{center}

\section{Experiment Details}

We used Adam \citep{kingma2014adam} for learning the neural network
parameters with a learning rate of $10^{-4}$ and $10^{-3}$ for
the actor and critic respectively. For $Q$ we included $L_2$ weight
decay of $10^{-2}$ and used a discount factor of $\gamma = 0.99$. For
the soft target updates we used $\tau = 0.001$.  The neural networks
used the rectified non-linearity \citep{glorot2011deep} for all hidden
layers. The final output layer of the actor was a $\tanh$ layer, to bound
the actions. The low-dimensional networks had 2 hidden layers with 400 and
300 units respectively ($\approx$ 130,000 parameters).  Actions were
not included until the 2nd hidden layer of $Q$. When learning from
pixels we used 3 convolutional layers (no pooling) with 32 filters at
each layer. This was followed by two fully connected layers with 200
units ($\approx$ 430,000 parameters).
The final layer weights and biases
of both the actor and critic were initialized from a uniform
distribution $[-3 \times 10^{-3}, 3 \times 10^{-3}]$ and
$[3 \times 10^{-4}, 3 \times 10^{-4}]$ for the low dimensional and pixel
cases respectively. This was to ensure the initial outputs for the policy
and value estimates were near zero. The other layers were initialized from uniform
distributions $[-\frac{1}{\sqrt{f}}, \frac{1}{\sqrt{f}}]$ where $f$ is the fan-in
of the layer.
The actions were not included
until the fully-connected layers. We trained with minibatch sizes of 64
for the low dimensional problems and 16 on pixels. We used a replay buffer size
of $10^6$.

For the exploration noise process we used temporally correlated noise
in order to explore well in physical environments that have momentum.
We used an Ornstein–-Uhlenbeck process \citep{uhlenbeck1930theory}
with $\theta = 0.15$ and
$\sigma=0.2$. The Ornstein-–Uhlenbeck
process models the velocity of a Brownian particle with friction,
which results in temporally correlated values centered around 0.




\section{Planning algorithm}
Our planner is implemented as a model-predictive controller
\citep{tassa2012synthesis}: at every time step we run a single iteration of
trajectory optimization (using iLQG, \citep{todorov2005generalized}), starting
from the true state of the system.
Every single trajectory optimization is planned over a horizon between
250ms and 600ms, and this planning horizon recedes as the simulation of
the world unfolds, as is the case in model-predictive control.

The iLQG iteration begins with an initial rollout of the previous
policy, which determines the nominal trajectory.  We use repeated
samples of simulated dynamics to approximate a linear expansion of the
dynamics around every step of the trajectory, as well as a quadratic
expansion of the cost function.  We use this sequence of
locally-linear-quadratic models to integrate the value function
backwards in time along the nominal trajectory.  This \emph{back-pass}
results in a putative modification to the action sequence that will
decrease the total cost.  We perform a derivative-free line-search
over this direction in the space of action sequences by integrating
the dynamics forward (the \emph{forward-pass}), and choose the best
trajectory.  We store this action sequence in order to warm-start the
next iLQG iteration, and execute the first action in the simulator.
This results in a new state, which is used as the initial state in the
next iteration of trajectory optimization.
\section{Environment details}


\subsection{Torcs environment}

For the Torcs environment we used a reward function
which provides a positive reward at each step for the velocity
of the car projected along the track direction and a penalty
of $-1$ for collisions. Episodes were terminated
if progress was not made along the track after 500
frames.

\subsection{MuJoCo environments}

For physical control tasks we used reward functions which
provide feedback at every step. In all tasks, the reward contained a small
action cost. For all tasks that have a static goal state (e.g. pendulum
swingup and reaching) we provide a smoothly varying reward based
on distance to a goal state, and in some cases an additional positive
reward when within a small radius of the target state.
For grasping and manipulation tasks we used a reward with a
term which encourages movement towards the payload
and a second component which encourages moving the payload to the target.
In locomotion tasks we reward forward action and
penalize hard impacts to encourage smooth rather than \
hopping gaits \citep{schulman2015trust}. In addition,
we used a negative reward and early termination for falls
which were determined by simple threshholds on the height and
torso angle (in the case of walker2d).

Table \ref{table:dims} states the
dimensionality of the problems and below is a summary
of all the physics environments.

\begin{table}[h]
\begin{center}
  \begin{tabular}{ l | c | c | c | c }
\textbf{task name }&  $\bf \dim(s)$ & $\bf \dim(a)$
      & $\bf \dim(o)$ \\
      \hline
      blockworld1 & 18 & 5 & 43\\
blockworld3da & 31 & 9 & 102\\
canada & 22 & 7 & 62\\
canada2d & 14 & 3 & 29\\
cart & 2 & 1 & 3\\
cartpole & 4 & 1 & 14\\
cartpoleBalance & 4 & 1 & 14\\
cartpoleParallelDouble & 6 & 1 & 16\\
cartpoleParallelTriple & 8 & 1 & 23\\
cartpoleSerialDouble & 6 & 1 & 14\\
cartpoleSerialTriple & 8 & 1 & 23\\
cheetah & 18 & 6 & 17\\
fixedReacher & 10 & 3 & 23\\
fixedReacherDouble & 8 & 2 & 18\\
fixedReacherSingle & 6 & 1 & 13\\
gripper & 18 & 5 & 43\\
gripperRandom & 18 & 5 & 43\\
hardCheetah & 18 & 6 & 17\\
hardCheetahNice & 18 & 6 & 17\\
hopper & 14 & 4 & 14\\
hyq & 37 & 12 & 37\\
hyqKick & 37 & 12 & 37\\
movingGripper & 22 & 7 & 49\\
movingGripperRandom & 22 & 7 & 49\\
pendulum & 2 & 1 & 3\\
reacher & 10 & 3 & 23\\
reacher3daFixedTarget & 20 & 7 & 61\\
reacher3daRandomTarget & 20 & 7 & 61\\
reacherDouble & 6 & 1 & 13\\
reacherObstacle & 18 & 5 & 38\\
reacherSingle & 6 & 1 & 13\\
walker2d & 18 & 6 & 41\\

  \end{tabular}
  \caption{
    Dimensionality of the MuJoCo tasks: the dimensionality
    of the underlying physics model $\dim(s)$, number
    of action dimensions $\dim(a)$ and observation dimensions $\dim(o)$.
    } \label{table:dims}
\end{center}
\end{table}

\setlength{\extrarowheight}{10pt}
\begin{center}
  \begin{longtable}{ l | p{10cm} }
  \textbf{task name } &  \textbf{Brief Description} \\
      \hline
blockworld1                  & Agent is required to use an arm with gripper constrained to the 2D plane to grab a falling block and lift it against gravity to a fixed target position. \\
blockworld3da                & Agent is required to use a human-like arm with 7-DOF and a simple gripper to grab a block and lift it against gravity to a fixed target position. \\
canada                       & Agent is required to use a 7-DOF arm with hockey-stick like appendage to hit a ball to a target. \\
canada2d                     & Agent is required to use an arm with hockey-stick like appendage to hit a ball initialzed to a random start location to a random target location. \\
cart                         & Agent must move a simple mass to rest at 0.  The mass begins each trial in random positions and with random velocities. \\
cartpole                     & The classic cart-pole swing-up task.  Agent must balance a pole attached to a cart by applying forces to the cart alone.  The pole starts each episode hanging upside-down.   \\
cartpoleBalance              & The classic cart-pole balance task.  Agent must balance a pole attached to a cart by applying forces to the cart alone.  The pole starts in the upright positions at the beginning of each episode. \\
cartpoleParallelDouble       & Variant on the classic cart-pole.  Two poles, both attached to the cart, should be kept upright as much as possible. \\
cartpoleSerialDouble         & Variant on the classic cart-pole.  Two poles, one attached to the cart and the second attached to the end of the first, should be kept upright as much as possible. \\
cartpoleSerialTriple         & Variant on the classic cart-pole.  Three poles, one attached to the cart, the second attached to the end of the first, and the third attached to the end of the second, should be kept upright as much as possible. \\
cheetah                      & The agent should move forward as quickly as possible with a cheetah-like body that is constrained to the plane.  This environment is based very closely on the one introduced by \citet{wawrzynski2009real,wawrzynski2013autonomous}. \\
fixedReacher                 & Agent is required to move a 3-DOF arm to a fixed target position. \\
fixedReacherDouble           & Agent is required to move a 2-DOF arm to a fixed target position. \\
fixedReacherSingle           & Agent is required to move a simple 1-DOF arm to a fixed target position. \\
gripper                      & Agent must use an arm with gripper appendage to grasp an object and manuver the object to a fixed target.      \\
gripperRandom                & The same task as {\tt gripper} except that the arm object and target position are initialized in random locations. \\
hardCheetah                  & The agent should move forward as quickly as possible with a cheetah-like body that is constrained to the plane.  This environment is based very closely on the one introduced by \citet{wawrzynski2009real,wawrzynski2013autonomous}, but has been made much more difficult by removing the stabalizing joint stiffness from the model. \\
hopper                       & Agent must balance a multiple degree of freedom monoped to keep it from falling.      \\
hyq                          & Agent is required to keep a quadroped model based on the hyq robot from falling. \\
movingGripper                & Agent must use an arm with gripper attached to a moveable platform to grasp an object and move it to a fixed target. \\
movingGripperRandom          & The same as the movingGripper environment except that the object position, target position, and arm state are initialized randomly. \\
pendulum                     & The classic pendulum swing-up problem.  The pendulum should be brought to the upright position and balanced.  Torque limits prevent the agent from swinging the pendulum up directly. \\
reacher3daFixedTarget        & Agent is required to move a 7-DOF human-like arm to a fixed target position.      \\
reacher3daRandomTarget       & Agent is required to move a 7-DOF human-like arm from random starting locations to random target positions. \\
reacher                      & Agent is required to move a 3-DOF arm from random starting locations to random target positions. \\
reacherSingle                & Agent is required to move a simple 1-DOF arm from random starting locations to random target positions. \\
reacherObstacle              & Agent is required to move a 5-DOF arm around an obstacle to a randomized target position. \\
walker2d                     & Agent should move forward as quickly as possible with a bipedal walker constrained to the plane without falling down or pitching the torso too far forward or backward. \\
\end{longtable}
 \end{center}

\end{document}